\documentclass[11pt,a4paper]{article}

\pdfoutput=1

\usepackage{a4wide}
\usepackage{graphicx}
\usepackage[usenames]{color,colortbl}
\usepackage{pstricks,pst-node,pst-plot}
\usepackage{verbatim}
\usepackage{placeins}
\usepackage{array}
\usepackage{amstext,amsmath,amssymb}
\usepackage{booktabs}
\usepackage{lscape}
\usepackage{rotating}
\usepackage{subfigure}
\usepackage{algorithm}
\usepackage{algorithmic}
\usepackage{url}
\usepackage{multirow}
\usepackage{empheq,fancybox}
\usepackage{amsfonts}
\usepackage{rotating}
\usepackage{tabularx}

\definecolor{lightgray}{gray}{.85}
\newcommand{\ccg}{\cellcolor{lightgray}}

\newcommand{\Spartial}{\mathcal{S}_{\mbox{\tiny partial}}}

\newcommand{\MIPorig}{\mbox{\textsc{Ilp}}_{\mbox{\tiny orig}}}
\newcommand{\MIPphone}{\mbox{\textsc{Ilp}}_{\mbox{\tiny ph1}}}
\newcommand{\MIPphtwo}{\mbox{\textsc{Ilp}}_{\mbox{\tiny ph2}}}

\begin{document}

\title{Mathematical Programming Strategies for Solving the Minimum Common String Partition Problem}

\author{Christian Blum$^{1,2}$, Jos{\'e} A.~Lozano$^1$, and Pedro Pinacho Davidson$^{1,3}$  \\
~\\
$^1$Department of Computer Science and Artificial Intelligence\\ 
University of the Basque Country UPV/EHU, San Sebastian, Spain \\
{\sf \{christian.blum,ja.lozano\}@ehu.es}\\
~\\
$^2$IKERBASQUE\\
Basque Foundation for Science, Bilbao, Spain\\
~\\
$^3$Escuela de Inform{\'a}tica\\
Universidad Santo Tom{\'a}s, Concepci{\'o}n, Chile\\
{\sf ppinacho@santotomas.cl}}

\date{}

\maketitle

\begin{abstract}
The minimum common string partition problem is an NP-hard combinatorial optimization problem with applications in computational biology. In this work we propose the first integer linear programming model for solving this problem. Moreover, on the basis of the integer linear programming model we develop a deterministic 2-phase heuristic which is applicable to larger problem instances. The results show that provenly optimal solutions can be obtained for problem instances of small and medium size from the literature by solving the proposed integer linear programming model with CPLEX. Furthermore, new best-known solutions are obtained for all considered problem instances from the literature. Concerning the heuristic, we were able to show that it outperforms heuristic competitors from the related literature.
\end{abstract}

\section{Introduction}
\label{sec:intro}

Optimization problems related to strings---such as protein or DNA sequences---are very common in bioinformatics. Examples include string consensus problems such as the far-from most string problem~\cite{MouBabMon12a,MenOliPar05}, the longest common subsequence problem and its variants~\cite{HsuDu84lcs,SmiWat81:jmb}, and alignment problems~\cite{Gus97:sequence-algorithms}. These problems are often computationally very hard, if not even $NP$-hard~\cite{GJ79}. In this work we deal with the \emph{minimum common string partition} (MCSP) problem, which can be described as follows. We are given two related input strings that have to be partitioned each into the same collection of substrings. The size of the collection is subject to minimization. A formal description of the problem will be provided in Section~\ref{sec:problem-description}. The MCSP problem has applications, for example, in the bioinformatics field. Chen et al.~\cite{CheEtAl05:pacific} point out that the MCSP problem is closely related to the problem of sorting by reversals with duplicates, a key problem in genome rearrangement. 

In this paper we introduce the first integer linear program (ILP) for solving the MCSP problem. An experimental evaluation on problem instances from the related literature shows that this ILP can be efficiently solved, for example, by using any version of IBM ILOG CPLEX. However, a study on new instances of larger size demonstrates the limitations of the model. Therefore, we additionally introduce a deterministic 2-phase heuristic which is strongly based on the original ILP. The experimental evaluation shows that the heuristic is applicable to larger problem instances than the original ILP. Moreover, it is shown that the heuristic outperforms competitor algorithms from the related literature on known problem instances.

\subsection{Problem Description}
\label{sec:problem-description}

The MCSP problem can technically be described as follows. Given are two input strings $s_1$ and $s_2$, both of length $n$ over a finite alphabet $\Sigma$. These two strings are required to be \emph{related}, which means that each letter appears the same number of times in each of them. Note that this definition implies that $s_1$ and $s_2$ have the same length. A valid solution to the MCSP problem is obtained by partitioning $s_1$ into a set $P_1$ of non-overlapping substrings, and $s_2$ into a set $P_2$ of non-overlapping substrings, such that $P_1 = P_2$. Moreover, we are interested in finding a valid solution such that $|P_1| = |P_2|$ is minimal.

Consider the following example. Given are DNA sequences $s_1 = \mbox{\bf AGACTG}$ and $s_2 = \mbox{\bf ACTAGG}$. Obviously, $s_1$ and $s_2$ are related because {\bf A} and ${\bf G}$ appear twice in both input strings, while {\bf C} and {\bf T} appear once. A trivial valid solution can be obtained by partitioning both strings into substrings of length 1, that is, $P_1 = P_2 = \{\mbox{\bf A}, \mbox{\bf A}, \mbox{\bf C}, \mbox{\bf T}, \mbox{\bf G}, \mbox{\bf G}\}$. The objective function value of this solution is 6. However, the optimal solution, with objective function value 3, is $P_1 = P_2 = \{\mbox{\bf ACT}, \mbox{\bf AG}, \mbox{\bf G}\}$.

\subsection{Related Work}

The MCSP problem has been introduced by Chen et al. \cite{CheEtAl05:pacific} due to its relation to genome rearrangement. More specifically, it has applications in biological questions such as: May a given DNA string possibly be obtained by rearrangements of another DNA string? The general problem has been shown to be $NP$-hard even in very restrictive cases~\cite{Goldstein2004}. Other papers concerning problem hardness consider, for example, the $k$-MCSP problem, which is the version of the MCSP problem in which each letter occurs at most $k$ times in each input string. The 2-MCSP problem was shown to be APX-hard in~\cite{Goldstein2004}. When the input strings are over an alphabet of size $c$, the corresponding problem is denoted as MCSP$^c$. Jiang et al.~proved that the decision version of the MCSP$^c$ problem is $NP$-complete when $c \geq 2$~\cite{Jiang2010}.

The MCSP has been considered quite extensively by researchers dealing with the approximability of the problem. Cormode and Muthukrishnan~\cite{Cormode2002}, for example, proposed an $O(lognlog^*n)$-approximation for the \emph{edit distance with moves} problem, which is a more general case of the MCSP problem. Shapira and Storer~\cite{Shapira2002} extended on this result. Other approximation approaches for the MCSP problem have been proposed in~\cite{Kolman2007}. In this context, Chrobak et al.~\cite{Chrobak2004} studied a simple greedy approach for the MCSP problem, showing that the approximation ratio concerning the 2-MCSP problem is 3, and for the 4-MCSP problem the approximation ratio is $\Omega(log(n))$. In the case of the general MCSP problem, the approximation ratio is between $\Omega(n^{0.43})$ and $O(n^{0.67})$, assuming that the input strings use an alphabet of size $O(log(n))$. Later Kaplan and Shafir~\cite{Kaplan2006} raised the lower bound to~$\Omega(n^{0.46})$. Kolman proposed a modified version of the simple greedy algorithm with an approximation ratio of $O(k^2)$ for the $k$-MCSP~\cite{Kolman2005}. Recently, Goldstein and Lewenstein proposed a greedy algorithm for the MCSP problem that runs in $O(n)$ time (see~\cite{Goldstein2011}). He~\cite{He2007} introduced a greedy algorithm with the aim of obtaining better average results.

Damaschke~\cite{raey} was the first one to study the fixed-parameter tractability (FPT) of the problem. Later, Jiang et al.~\cite{Jiang2010} showed that both the $k$-MCSP and MCSP$^c$ problems admit FPT algorithms when $k$ and $c$ are constant parameters. Finally, Fu et al.~\cite{BinFu2011} proposed a $O(2^nn^{O(1)})$ time algorithm for the general case and an $O(n(log n)^2)$ time algorithm applicable under some constraints. 

To our knowledge, the only metaheuristic approaches that have been proposed in the related literature for the MCSP problem are (1) the {${\cal MAX}$}-{${\cal MIN}$} {A}nt {S}ystem by Ferdous and Sohel~\cite{Ferdous2013,FerRah14:arxiv} and (2) the probabilistic tree search algorithm by Blum et al.~\cite{BluEtAl14:hm}. Both works applied their algorithm to a range of artificial and real DNA instances from~\cite{Ferdous2013}.

\subsection{Organization of the Paper}

The remaining part of the paper is organized as follows. In Section~\ref{sec:mip}, the ILP model for solving the MCSP is outlined. Moreover, an experimental evaluation is provided. The deterministic heuristic, together with an experimental evaluation, is described in Section~\ref{sec:heuristic}. Finally, in Section~\ref{sec:conclusions} we provide conclusions and an outlook to future work.

\section{An Integer Linear Program to Solve the MCSP}
\label{sec:mip}

In the following we present the first ILP model for solving the MCSP. For this, the definitions provided in the following are required. Note that an illustrative example is provided in Section~\ref{sec:example}.

\subsection{Preliminaries}
\label{sec:prelim}

Henceforth, a \emph{common block} $b_i$ of input strings $s_1$ and $s_2$ is denoted as a triple $(t_i,k1_i,k2_i)$ where $t_i$ is a string which can be found starting at position $1 \leq k1_i \leq n$ in string $s_1$ and starting at position $1 \leq k2_i \leq n$ in string $s_2$. Moreover, let $B = \{b_1, \ldots, b_m\}$ be the (ordered) set of all possible common blocks of $s_1$ and $s_2$.\footnote{The way in which $B$ is ordered is of no importance.} Given the definition of $B$, any valid solution $\mathcal{S}$ to the MCSP problem is a subset of $B$---that is, $\mathcal{S} \subset B$---such that:
\begin{enumerate}
  \item $\sum_{b_i \in \mathcal{S}} |t_i| = n$, that is, the sum of the length of the strings corresponding to the common blocks in $\mathcal{S}$ is equal to the length of the input strings.
  \item For any two common blocks $b_i, b_j \in \mathcal{S}$ it holds that their corresponding strings neither overlap in $s_1$ nor in $s_2$.
\end{enumerate}
Moreover, any (valid) partial solution $\Spartial$ is a subset of $B$ fulfilling the following conditions: (1) $\sum_{b_i \in \mathcal{\Spartial}} |t_i| < n$ and (2) for any two common blocks $b_i, b_j \in \Spartial$ it holds that their corresponding strings neither overlap in $s_1$ nor in $s_2$. Note that any valid partial solution can be extended to be a valid solution. Furthermore, given a partial solution $\Spartial$, set $B(\Spartial) \subset B$ denotes the set of common blocks that may be used in order to extend $\Spartial$ such that the result is again a valid (partial) solution. 

\subsection{The Integer Linear Program}
\label{sec:puremip}

First, two binary $m \times n$ matrices $M1$ and $M2$ are defined as follows. In both matrices, row $1 \leq i \leq m$ corresponds to common block $b_i \in B$. Moreover, a column $1 \leq j \leq n$ corresponds to position $j$ in input string $s_1$, respectively $s_2$. In general, the entries of matrix $M1$ are set to zero. However, in each row $i$, the positions that string $t_i$ (of common block $b_i$) occupies in input string $s_1$ are set to one. Correspondingly, the entries of matrix $M2$ are set to zero, apart from the fact that in each row $i$ the positions occupied by string $t_i$ in input string $s_2$ are set to one. Henceforth, the position $(i,j)$ of a matrix $M$ is denoted by $M_{i,j}$. Finally, we introduce for each common block $b_i \in B$ a binary variable $x_i$. With these definitions we can express the MCSP in form of the following integer linear program, henceforth referred to by $\MIPorig$.

\begin{empheq}[box=\shadowbox*]{align}
  & \mbox{\bf min} \sum_{i=1}^{m} x_i \\
  \shoveright{\text{\bf subject to:}} & \nonumber \\
  \sum_{i=1}^m |t_i| \cdot x_i      & = n   \label{eqn:const1} \\
  \sum_{i=1}^m M1_{i,j} \cdot x_i    & = 1   \;\;\; \mbox{for } j=1,\ldots,n \label{eqn:const2} \\
  \sum_{i=1}^m M2_{i,j} \cdot x_i    & = 1   \;\;\; \mbox{for } j=1,\ldots,n \label{eqn:const3} \\
  x_i          & \in \{0, 1\} \;\;\; \mbox{for } i=1,\ldots,m\nonumber
\end{empheq}

Hereby, the objective function minimizes the number of selected common blocks. Constraint (\ref{eqn:const1}) ensures that the sum of the length of the strings corresponding to the selected common blocks is equal to $n$. Finally, constraints (\ref{eqn:const2}) make sure that the strings corresponding to the selected common blocks do not overlap in input string $s_1$, while constraints (\ref{eqn:const3}) make sure that the strings corresponding to the selected common blocks do not overlap in input string $s_2$.

\subsection{Example}
\label{sec:example}

As an example, consider the small problem instance from Section~\ref{sec:problem-description}. The complete set of common blocks ($B$) as induced by input strings $s_1 = \mbox{\bf AGACTG}$ and $s_2 = \mbox{\bf ACTAGG}$ is as follows:
\begin{equation}
B = \begin{Bmatrix*}[l]
b_1 = $(\mbox{\bf ACT}, 3, 1)$ \\
b_2 = $(\mbox{\bf AG}, 1, 4)$ \\
b_3 = $(\mbox{\bf AC}, 3, 1)$ \\
b_4 = $(\mbox{\bf CT}, 4, 2)$ \\
b_5 = $(\mbox{\bf A}, 1, 1)$ \\
b_6 = $(\mbox{\bf A}, 1, 4)$ \\
b_7 = $(\mbox{\bf A}, 3, 1)$ \\
b_8 = $(\mbox{\bf A}, 3, 4)$ \\
b_9 = $(\mbox{\bf C}, 4, 2)$ \\
b_{10} = $(\mbox{\bf T}, 5, 3)$ \\
b_{11} = $(\mbox{\bf G}, 2, 5)$ \\
b_{12} = $(\mbox{\bf G}, 2, 6)$ \\
b_{13} = $(\mbox{\bf G}, 6, 5)$ \\
b_{14} = $(\mbox{\bf G}, 6, 6)$
\end{Bmatrix*} \nonumber
\end{equation}

\noindent Given set $B$, matrices $M1$ and $M2$ are the following ones:

\begin{center}
\begin{tabular}{ccc}
$M1 = \begin{pmatrix}
  0 & 0 & 1 & 1 & 1 & 0 \\
  1 & 1 & 0 & 0 & 0 & 0 \\
  0 & 0 & 1 & 1 & 0 & 0 \\
  0 & 0 & 0 & 1 & 1 & 0 \\
  1 & 0 & 0 & 0 & 0 & 0 \\
  1 & 0 & 0 & 0 & 0 & 0 \\
  0 & 0 & 1 & 0 & 0 & 0 \\
  0 & 0 & 1 & 0 & 0 & 0 \\
  0 & 0 & 0 & 1 & 0 & 0 \\
  0 & 0 & 0 & 0 & 1 & 0 \\
  0 & 1 & 0 & 0 & 0 & 0 \\
  0 & 1 & 0 & 0 & 0 & 0 \\
  0 & 0 & 0 & 0 & 0 & 1 \\
  0 & 0 & 0 & 0 & 0 & 1 
\end{pmatrix}$
&
$\;\;\;\;$
&
$M2 = \begin{pmatrix}
  1 & 1 & 1 & 0 & 0 & 0 \\
  0 & 0 & 0 & 1 & 1 & 0 \\
  1 & 1 & 0 & 0 & 0 & 0 \\
  0 & 1 & 1 & 0 & 0 & 0 \\
  1 & 0 & 0 & 0 & 0 & 0 \\
  0 & 0 & 0 & 1 & 0 & 0 \\
  1 & 0 & 0 & 0 & 0 & 0 \\
  0 & 0 & 0 & 1 & 0 & 0 \\
  0 & 1 & 0 & 0 & 0 & 0 \\
  0 & 0 & 1 & 0 & 0 & 0 \\
  0 & 0 & 0 & 0 & 1 & 0 \\
  0 & 0 & 0 & 0 & 0 & 1 \\
  0 & 0 & 0 & 0 & 1 & 0 \\
  0 & 0 & 0 & 0 & 0 & 1 
\end{pmatrix}$
\end{tabular}
\end{center}

The optimal solution to this instance is $\mathcal{S} = \{b_1, b_2, b_{14}\}$. It can easily be verified that this solution respects constraints (2-4) of the ILP model.

\subsection{Experimental Evaluation}
\label{sec:mip-exp}

In the following we will provide an experimental evaluation of model $\MIPorig$. The model was implemented in ANSI C++ using GCC 4.7.3 for compiling the software. Moreover, the model was solved with IBM ILOG CPLEX V12.1. The experimental results that we outline in the following were obtained on a cluster of PCs with "Intel(R) Xeon(R) CPU 5130" CPUs of 4 nuclei of 2000 MHz and 4 Gigabyte of RAM.

\subsubsection{Problem Instances}

For testing model $\MIPorig$ we chose the same set of benchmark instances that was used by Ferdous and Sohel in~\cite{Ferdous2013} for the experimental evaluation of their ant colony optimization approach. This set contains, in total, 30 artificial instances and 15 real-life instances consisting of DNA sequences. Remember, in this context, that each problem instance consists of two related input strings. Moreover, the benchmark set consists of four subsets of instances. The first subset (henceforth labelled \textsc{Group1}) consists of 10 artificial instances in which the input strings are maximally of length 200. The second set (\textsc{Group2}) consists of 10 artificial instances with input string lengths between 201 and 400. In the third set (\textsc{Group3}) the input strings of the 10 artificial instances have lengths between 401 and 600. Finally, the fourth set (\textsc{Real}) consists of 15 real-life instances of various lengths.

\subsubsection{Results}
\label{sec:mip:results}

The results are shown in Tables~\ref{tab:results:group1}-\ref{tab:results:real}, in terms of one table per instance set. The structure of these tables is as follows. The first column provides the instance identifiers. The second column contains the results of the greedy algorithm from~\cite{Chrobak2004} (results were taken from~\cite{Ferdous2013}). The third column provides the value of the best solution found in four independent runs per problem instance (with a CPU time limit of 7200 seconds per run) by the \textsc{Aco} approach by Ferdous and Sohel~\cite{Ferdous2013,FerRah14:arxiv}.\footnote{In this context, note that the experiments for \textsc{Aco} were performed on a computer with an "Intel(R) 2 Quad" CPU with 2.33 GHz and 4 GB of RAM.} The fourth column provides the value of the best solution found in 10 independent runs per problem instance (with a CPU time limit of 1000 seconds per run) by the probabilistic tree search algorithm (henceforth labelled \textsc{TreSea}) by Blum et al.~\cite{BluEtAl14:hm}. \textsc{TreSea} was run on the same machines as the ones used for the current work. Finally, the last four table columns are dedicated to the presentation of the results provided by solving model $\MIPorig$. The first one of these columns provides the value of the best solution found within 3600 CPU seconds. In case the optimality of the corresponding solution was proved by CPLEX, the value is marked by an asterix. The second column dedicated to $\MIPorig$ provides the computation time (in seconds). In case of having solved the corresponding problem to optimality, this column only displays one value indicating the time needed by CPLEX to solve the problem. Otherwise, this column provides two values in the form X/Y, where X corresponds to the time at which CPLEX was able to find the first valid solution, and Y corresponds to the time at which CPLEX found the best solution within 3600 CPU seconds. The third one of the columns dedicated to $\MIPorig$ shows the optimality gap, which refers to the gap between the value of the best valid solution and the current lower bound at the time of stopping a run. Finally, the last column indicates the size of set $B$, that is, the size of the complete set of common blocks. Note that this value corresponds to the number of variables used by $\MIPorig$. The best result (among all algorithms) for each problem instance is marked by a grey background, and the last row of each table provides averages over the whole table. \\

\begin{table}[t!]
\caption{Results for the 10 instances of \textsc{Group1}.}
\label{tab:results:group1}
\centering
\scalebox{0.8}{
\begin{tabular}{rrrrrrrrrrrr} \hline\hline
{\bf id} & $\;\;\;$ & \textsc{Greedy} & $\;\;\;$ & \textsc{Aco} & $\;\;\;$ & \textsc{TreSea} & $\;\;\;$ & \multicolumn{4}{c}{$\MIPorig$} \\ \cline{3-3} \cline{5-5} \cline{7-7} \cline{9-12}
                     &          & {\bf value}     &          &  {\bf best}  &          & {\bf best}          &          & {\bf value} & {\bf time (s)} & {\bf gap} & {\bf $|B|$} \\ \hline
1          && 46   && 42   && 42   && $^*$\ccg41      &       1       &       0.0\%    & 4299  \\
2          && 56   && 51   && 48   && $^*$\ccg47      &       3       &       0.0\%    & 6211  \\
3          && 62   && 55   && 56   && $^*$\ccg52      &       30      &       0.0\%    & 8439  \\
4          && 46   && 43   && 43   && $^*$\ccg41      &       2       &       0.0\%    & 4299  \\
5          && 44   && 43   && 41   && $^*$\ccg40      &       1       &       0.0\%    & 4718  \\
6          && 48   && 42   && 41   && $^*$\ccg40      &       3       &       0.0\%    & 4435  \\
7          && 65   && 60   && 60   && $^*$\ccg55      &       38      &       0.0\%    & 8687  \\
8          && 51   && 47   && 45   && $^*$\ccg43      &       3       &       0.0\%    & 4995  \\
9          && 46   && 45   && 43   && $^*$\ccg42      &       2       &       0.0\%    & 4995  \\
10         && 63   && 59   && 58   && $^*$\ccg54      &       50      &       0.0\%    & 9699  \\ \hline
{\bf avg.} && 52.7 && 48.7 && 47.7 && \ccg45.5        &       13.3    &       0.0\%    & 6029.3  \\
\hline\hline
\end{tabular}}
\end{table}

\begin{table}[t!]
\caption{Results for the 10 instances of \textsc{Group2}.}
\label{tab:results:group2}
\centering
\scalebox{0.8}{
\begin{tabular}{rrrrrrrrrrrr} \hline\hline
{\bf id} & $\;\;\;$ & \textsc{Greedy} & $\;\;\;$ & \textsc{Aco} & $\;\;\;$ & \textsc{TreSea} & $\;\;\;$ & \multicolumn{4}{c}{$\MIPorig$} \\ \cline{3-3} \cline{5-5} \cline{7-7} \cline{9-12}
                     &          & {\bf value}     &          &  {\bf best}  &          & {\bf best}          &          & {\bf value} & {\bf time (s)} & {\bf gap} & {\bf $|B|$} \\ \hline
1          && 119   && 113   && 111   && \ccg98	  &	66/1969	    &	2.9\% & 37743\\
2          && 122   && 118   && 114   && \ccg106  &	129/1032    &	7.5\% &	47174\\
3          && 114   && 111   && 107   && \ccg97	  &	55/1216	    &	2.7\% &	36979\\
4          && 116   && 115   && 111   && \ccg102  & 	63/949	    &	4.9\% &	40960\\
5          && 135   && 132   && 127   && \ccg116  &	146/3299    &	6.7\% &	52697\\
6          && 108   && 105   && 102   && \ccg93	  &	56/1419	    &	5.6\% &	35650\\
7          && 108   && 98    && 96    && \ccg88	  &	41/2776	    &	6.0\% &	30839\\
8          && 123   && 118   && 114   && \ccg104  &	101/2980    &	5.1\% &	42668\\
9          && 124   && 119   && 113   && \ccg104  &	81/1630	    &	5.2\% &	42998\\
10         && 105   && 101   && 98    && \ccg89	  &	32/1458	    &	3.6\% &	31169\\ \hline
{\bf avg.} && 117.4 && 113.0 && 109.3 && \ccg99.7 &     77/1873     &   5.0\% & 39887.7 \\
\hline\hline
\end{tabular}}
\end{table}

\begin{table}[t!]
\caption{Results for the 10 instances of \textsc{Group3}.}
\label{tab:results:group3}
\centering
\scalebox{0.8}{
\begin{tabular}{rrrrrrrrrrrrrr} \hline\hline
{\bf id} & $\;\;\;$ & \textsc{Greedy} & $\;\;\;$ & \textsc{Aco} & $\;\;\;$ & \textsc{TreSea} & $\;\;\;$ & \multicolumn{4}{c}{$\MIPorig$} \\ \cline{3-3} \cline{5-5} \cline{7-7} \cline{9-12}
                     &          & {\bf value}     &          &  {\bf best}  &          & {\bf best}          &          & {\bf value} & {\bf time (s)} & {\bf gap} & {\bf $|B|$} \\ \hline
1          && 182   && 177   && 171   && \ccg155   &	 733/1398 &	 7.5\%  & 110973   \\
2          && 175   && 175   && 168   && \ccg155   &	 553/869  &	 7.7\%  & 102670   \\
3          && 196   && 187   && 185   && \ccg166   &	 746/2183 &	 8.5\%  & 119287   \\
4          && 192   && 184   && 179   && \ccg159   &	 731/1200 &	 6.9\%  & 114975   \\
5          && 176   && 171   && 163   && \ccg150   &	 485/886  &	 9.7\%  & 99775    \\
6          && 170   && 160   && 162   && \ccg147   &	 399/764  &	 9.1\%  & 88839    \\
7          && 173   && 167   && 161   && \ccg149   &	 524/990  &	 9.8\%  & 95765    \\
8          && 185   && 175   && 169   && \ccg151   &	 492/3584 &	 6.7\%  & 97400    \\
9          && 174   && 172   && 169   && \ccg158   &	 571/1186 &	 10.9\% & 104186   \\
10         && 171   && 167   && 161   && \ccg148   &	 547/1446 &	 9.1\%  & 98237    \\ \hline
{\bf avg.} && 179.4 && 173.5 && 168.8 && \ccg153.8 &     578/1451 &      8.6\%  & 103211.0 \\
\hline\hline
\end{tabular}}
\end{table}

\begin{table}[t!]
\caption{Results for the 15 instances of set \textsc{Real}.}
\label{tab:results:real}
\centering
\scalebox{0.8}{
\begin{tabular}{rrrrrrrrrrrr} \hline\hline
{\bf id} & $\;\;\;$ & \textsc{Greedy} & $\;\;\;$ & \textsc{Aco} & $\;\;\;$ & \textsc{TreSea} & $\;\;\;$ & \multicolumn{4}{c}{$\MIPorig$} \\ \cline{3-3} \cline{5-5} \cline{7-7} \cline{9-12}
                     &          & {\bf value}     &          &  {\bf best}  &          & {\bf best}          &          & {\bf value} & {\bf time (s)} & {\bf gap} & {\bf $|B|$} \\ \hline
1          && 95    && 87    && 86    && $^*$\ccg78 &	   972      &	   0.0\% & 22799 \\
2          && 161   && 155   && 154   && \ccg139   &	   432/752  &	   9.2\% & 80523	\\
3          && 121   && 116   && 113   && \ccg104   &	   125/3580 &	   5.6\% & 45869	\\
4          && 173   && 164   && 158   && \ccg144   &	   577/1730 &	   6.5\% & 91663	\\
5          && 172   && 171   && 165   && \ccg150   &	   778/2509 &	   7.9\% & 108866	\\
6          && 153   && 145   && 143   && \ccg128   &	   257/3578 &	   6.5\% & 70655	\\
7          && 140   && 140   && 131   && \ccg121   &	   359/2187 &	   6.9\% & 73502	\\
8          && 134   && 130   && 128   && \ccg116   &	   275/3365 &	   6.8\% & 65560	\\
9          && 149   && 146   && 142   && \ccg131   &	   399/613  &	   8.8\% & 75833	\\
10         && 151   && 148   && 144   && \ccg131   &	   311/1771 &	   7.2\% & 69560	\\
11         && 126   && 124   && 121   && \ccg110   &	   205/3711 &	   4.8\% & 56160	\\
12         && 143   && 137   && 138   && \ccg126   &	   299/793  &	   9.8\% & 70861	\\
13         && 180   && 180   && 171   && \ccg156   &	   784/1130 &	   7.1\% & 115810	\\
14         && 152   && 147   && 146   && \ccg134   &	   370/2456 &	   9.7\% & 73449	\\
15         && 157   && 160   && 152   && \ccg139   &	   560/1762 &	   7.7\% & 91060	\\ \hline
{\bf avg.} && 147.1 && 143.3 && 139.5 && \ccg127.1 &       409/2131 &      7.0\% & 74163.9     \\
\hline\hline
\end{tabular}}
\end{table}

The following conclusions can be drawn when analyzing the results. First, CPLEX is able to solve all instances of \textsc{Group1} to optimality. This is done, on average, in about 13 seconds. Moreover, none of the existing algorithms was able to find any of these optimal solutions. Second, CPLEX was also able to find new best-known solutions for all remaining 35 problem instances, even though it was not able to prove optimality within 3600 CPU seconds, which is indicated by the positive optimality gaps. An exception is instance 1 of set \textsc{Real} which also could be solved to optimality. Third, the improvements over the competitor algorithms obtained by solving $\MIPorig$ with CPLEX are remarkable. In particular, the average improvement (in percent) over \textsc{TreSea}, the best competitor from the literature, is $4.8\%$ in the case of \textsc{Group1}, $9.2\%$ in the case of \textsc{Group2}, $9.7\%$ in the case of \textsc{Group3}, and $9.8\%$ in the case of \textsc{Real}. \\

In order to study the limits of solving $\MIPorig$ with CPLEX we randomly generated larger DNA instances. In particular, we generated one random instance for each input string size from $\{800, 1000, 1200, 1400, 1600, 1800, 2000\}$. CPLEX was stopped when at least 3600 CPU seconds had passed and at least one feasible solution had been found. However, if after 12 CPU hours still no feasible solution was found, the execution was stopped as well. The results are shown in Table~\ref{tab:results:mip:large}. The first column of this table provides the length of the corresponding random instance. The remaining four columns contain the same information as already explained in the context of Tables~\ref{tab:results:group1}-\ref{tab:results:real}, just that column {\bf time (s)} simply provides the computation time (in seconds) at which the best solution was found. Analyzing the results we can observe that the application of CPLEX to $\MIPorig$ quickly becomes unpractical with growing input string size. For example, the first valid solution for the instance with string length 1400 was found after 20616 seconds. Concerning the largest problem instance, no valid solution was found within 12 CPU hours.

\begin{table}[!t]
\caption{Results of applying CPLEX to $\MIPorig$ in the context of larger instances.}
\label{tab:results:mip:large}
\centering
\scalebox{0.8}{
\begin{tabular}{rrrrr} \hline\hline
{\bf length} & {\bf value} & {\bf time (s)} & {\bf gap} & {\bf $|B|$} \\ \hline
800  & 210  &  3228  &   $10.7\%$ &  214622  \\
1000 & 304  &  2922  &   $26.4\%$ &  334411  \\
1200 & 342  &  6220  &   $22.6\%$ &  480908  \\
1400 & 401  &  12124 &   $24.9\%$ &  653401  \\
1600 & 442  &  20616 &   $24.1\%$ &  854500  \\
1800 & 486  &  37304 &   $24.0\%$ &  1084533 \\
2000 & n.a. &  n.a.  &   n.a. &  1335893 \\
\hline\hline
\end{tabular}}
\end{table}

\section{A MIP-Based Heuristic}
\label{sec:heuristic}

\begin{figure}[!t]
\centering
\subfigure[Instances of set \textsc{Group1}.]{
\label{fig:idea:heuristic:a}
\includegraphics[width=0.45\textwidth]{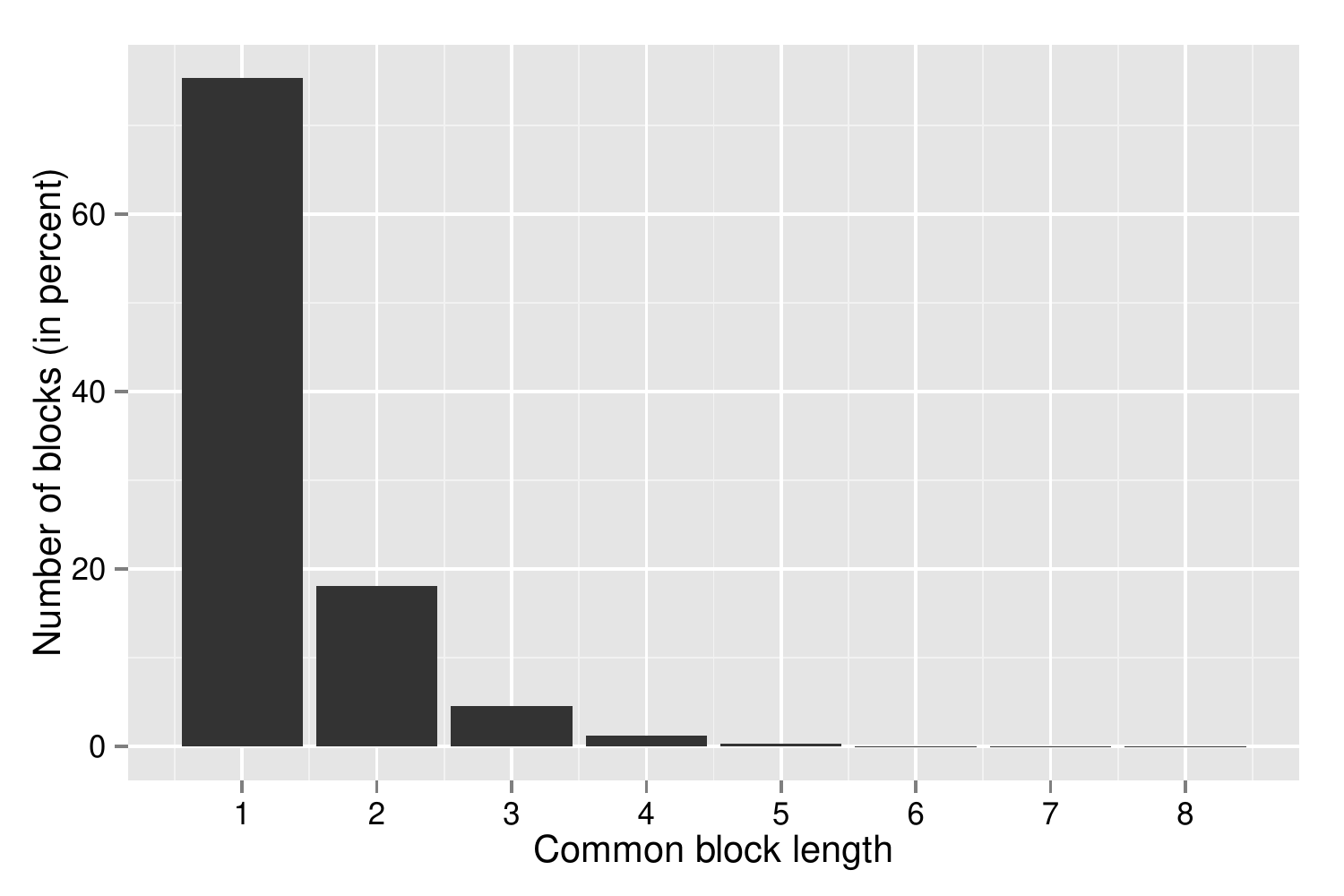}
}
\subfigure[Instances of set \textsc{Group2}.]{
\label{fig:idea:heuristic:b}
\includegraphics[width=0.45\textwidth]{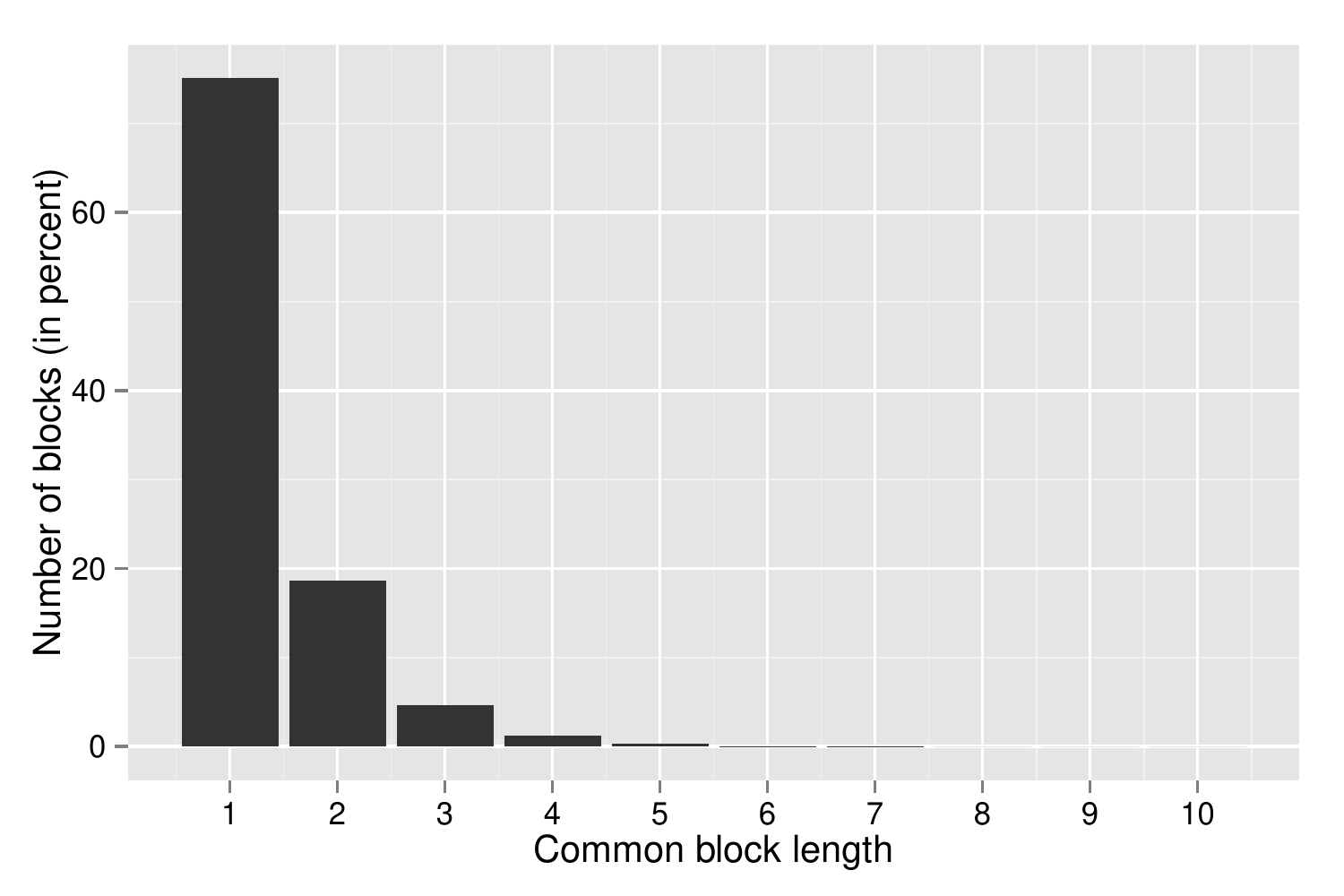}
}
\subfigure[Instances of set \textsc{Group3}.]{
\label{fig:idea:heuristic:c}
\includegraphics[width=0.45\textwidth]{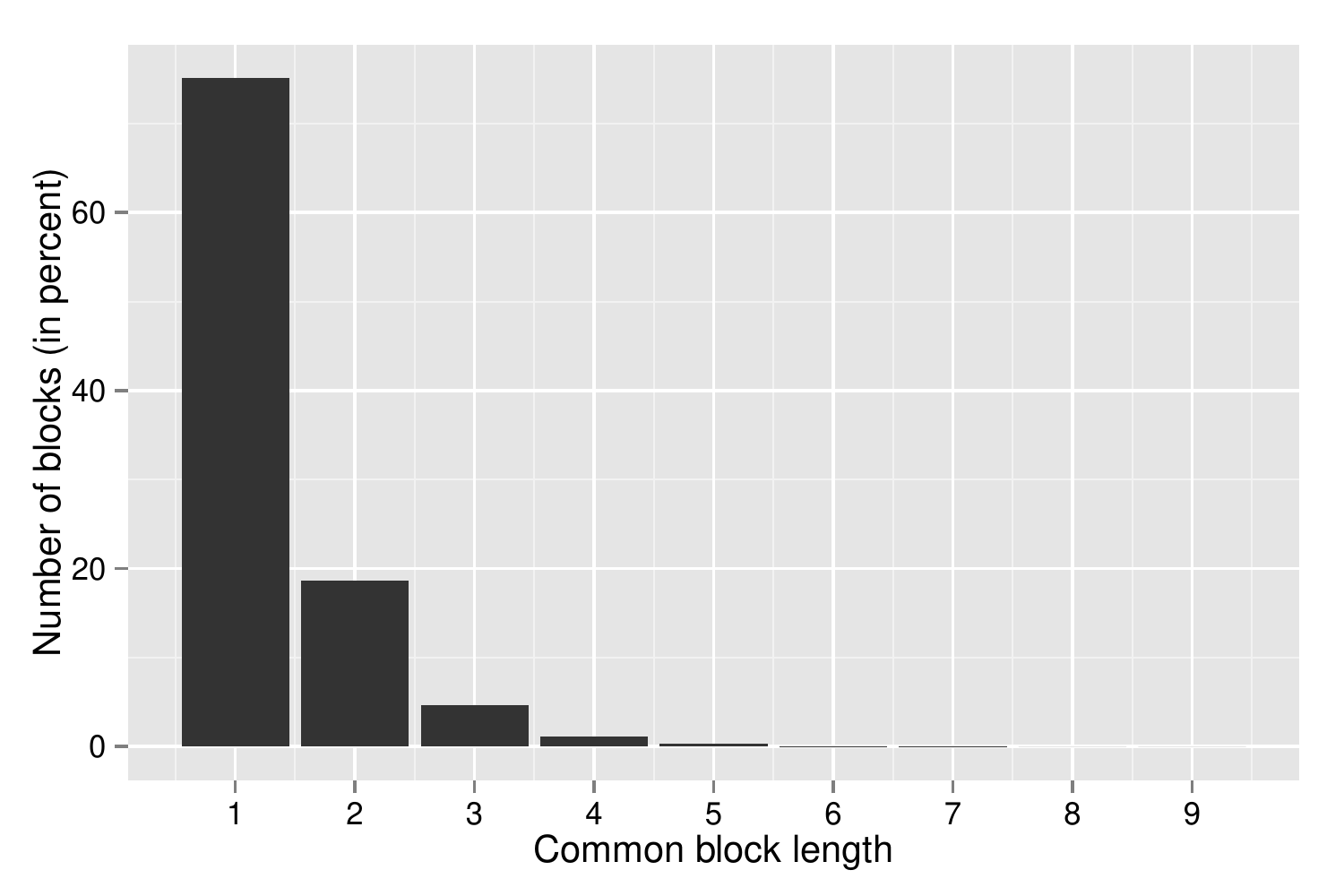}
}
\subfigure[Instances of set \textsc{Real}.]{
\label{fig:idea:heuristic:d}
\includegraphics[width=0.45\textwidth]{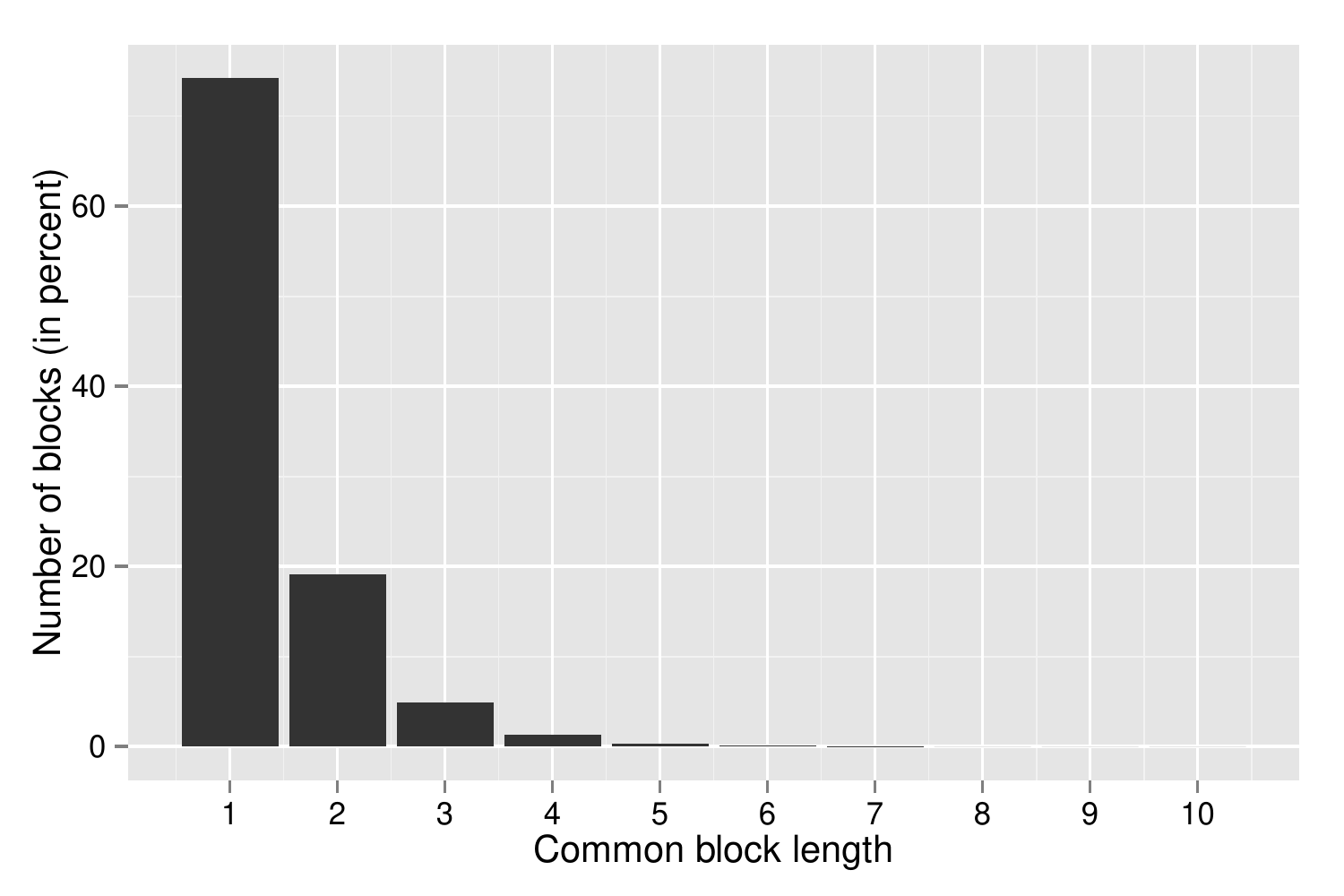}
}
\caption{Distribution of the string lengths corresponding to the complete set of common blocks. The distributions are shown averaged over all instances for each of the four sets of problem instances.}
\label{fig:idea:heuristic}
\end{figure}

As shown at the end of the previous section, the application of CPLEX to $\MIPorig$ reaches its limits starting from an input string size of about 1200. However, if it were possible to considerably reduce the size of the set of common blocks ($B$), mathematical programming might still be an option to obtain good (heuristic) solutions. With this idea in mind we studied the distribution of the lengths of the strings of the common blocks in $B$ for all 45 problem instances. This distribution is shown---averaged over the instances of each of the four instance sets--in Figure~\ref{fig:idea:heuristic}. Analyzing these distributions it can be observed, first of all, that the distribution does not seem to depend on instance size.\footnote{Most probably the distribution would change in some way when changing the size of the alphabet.} However, the important aspect to observe is that around $75\%$ of all the common blocks contain strings of length $1$. Moreover, only a very small portion of these common blocks will form part of an optimal solution. In comparison, it is reasonable to assume that a much larger percentage of the blocks corresponding to large strings will form part of an optimal solution. These observations gave rise to the heuristic which is outlined in the following.

\subsection{Heuristic}

The proposed heuristic works in two phases. In the first phase, a subset of $B$ (the complete set of common blocks) must be chosen. For this purpose, let $B_{\geq l}$ (where $l \geq 1$) denote the subset of $B$ that contains all common blocks $b_i$ from $B$ with $|t_i| \geq l$, that is, all blocks whose corresponding string is longer or equal than $l$. Note, in this context, that $B_{\geq 1} = B$. Moreover, note that $|B_{\geq 1}| \geq |B_{\geq 2}| \geq |B_{\geq 3}| \geq \ldots \geq |B_{\geq \infty}|$. Let $l_{\mbox{\tiny max}}$ be the smallest value for $l$ such that $|B_{\geq l_{\mbox{\tiny max}}}| > 0$. Observe that $B_{\geq l_{\mbox{\tiny max}}}$ only contains the common blocks with the longest strings. Having chosen a specific value for $l$ from $[2,l_{\mbox{\tiny max}}]$, the following ILP, henceforth referred to as $\MIPphone$, may be solved.

\begin{empheq}[box=\shadowbox*]{align}
  & \mbox{\bf max} \sum_{b_i \in B_{\geq l}} (C \cdot |t_i| - 1) \cdot x_i \\
  \shoveright{\text{\bf subject to:}} & \nonumber \\
  \sum_{b_i \in B_{\geq l}} |t_i| \cdot x_i      & \leq n   \label{eqn:const4} \\
  \sum_{b_i \in B_{\geq l}} M1_{i,j} \cdot x_i    & \leq 1   \;\;\; \mbox{for } j=1,\ldots,n \label{eqn:const5} \\
  \sum_{b_i \in B_{\geq l}} M2_{i,j} \cdot x_i    & \leq 1   \;\;\; \mbox{for } j=1,\ldots,n \label{eqn:const6} \\
  x_i          & \in \{0, 1\} \;\;\; \mbox{for } b_i \in B_{\geq l} \nonumber
\end{empheq}

$\MIPphone$ is based on a binary variable $x_i$ for each common block $b_i \in B_{\geq l}$. Moreover, matrices $M1$ and $M2$ are the same as the ones introduced in Section~\ref{sec:puremip}, that is, they are defined over the whole set $B$. The objective function basically maximizes the sum of the lengths of the strings corresponding to the chosen common blocks. However, the length of the string corresponding to each block is multiplied by a large-enough constant $C$, and the result is decremented by one. This has the following effect. In case of several solutions for which the sum of the lengths of the strings corresponding to the selected common blocks is equal, the program prefers the solution that reaches this sum with fewer common blocks. The constraints (\ref{eqn:const4}-\ref{eqn:const6}) are the same as in $\MIPorig$ (see Section~\ref{sec:puremip}), apart from the fact that all equality symbols are replaced by the $\leq$-symbol. In short, the idea of $\MIPphone$ is to produce a partial solution for the original MCSP that covers as much as possible of both input strings, while choosing as few common blocks as possible. 

Solving $\MIPphone$ will henceforth be referred to as \emph{phase 1} of the proposed heuristic. Let us denote by $\mathcal{S}_{ph1}$ the solution provided by phase 1.\footnote{Remember that solutions are subsets of $B$.} Due to the constraints of $\MIPphone$ this solution is a valid partial solution to the original MCSP problem. The idea of the second phase is then to produce the best complete solution possible that contains $\mathcal{S}_{ph1}$. This is done by solving the following ILP, henceforth referred to as $\MIPphtwo$.

\begin{empheq}[box=\shadowbox*]{align}
  & \mbox{\bf min} \sum_{b_i \in B_{ph2}} x_i \\
  \shoveright{\text{\bf subject to:}} & \nonumber \\
  \sum_{b_i \in B_{ph2}} |t_i| \cdot x_i        & = n   \label{eqn:const7} \\
  \sum_{b_i \in B_{ph2}} M1_{i,j} \cdot x_i    & = 1   \;\;\; \mbox{for } j=1,\ldots,n \label{eqn:const8} \\
  \sum_{b_i \in B_{ph2}} M2_{i,j} \cdot x_i    & = 1   \;\;\; \mbox{for } j=1,\ldots,n \label{eqn:const9} \\
  x_i                                           & = 1   \;\;\; \mbox{for } b_i \in \mathcal{S}_{ph1} \label{eqn:const10} \\
  x_i          & \in \{0, 1\} \;\;\; \mbox{for } b_i \in B_{ph2} \nonumber
\end{empheq}

Hereby, $B_{ph2} := B(\mathcal{S}_{ph1}) \subset B$ is the set of common blocks that may be added to $\mathcal{S}_{ph1}$ without violating any constraints.\footnote{See Section~\ref{sec:prelim} for the definition of $B(\cdot)$.} Note that model $\MIPphtwo$ is the same as model $\MIPorig$, just that $\MIPphtwo$ only considers common blocks from $B_{ph2}$ and that it forces any solution to contain all common blocks from $\mathcal{S}_{ph1}$; see constraints (\ref{eqn:const10}). This completes the description of the heuristic. 

\subsection{Experimental Evaluation}

Just like model $\MIPorig$, the heuristic was implemented in ANSI C++ using GCC 4.7.3 for compiling the software. The two ILP models were solved with IBM ILOG CPLEX V12.1, and the same machines as for the experimental evaluation of $\MIPorig$ were used for running the experiments. 

As mentioned before, the heuristic may be applied for any value of $l$ from the interval $[2,l_{\mbox{\tiny max}}]$. In fact, we applied the heuristic to each of the 45 problem instances from sets \textsc{Group1}, \textsc{Group2}, \textsc{Group3}, and \textsc{Real}, with all possible values for $l$. In order not to spend too much computation time the following stopping criterion was used for each call to CPLEX concerning any of the two involved ILP models. CPLEX was stopped (1) in case a provenly optimal solution was obtained or (2) in case at least 50 CPU seconds were spent and the first valid solution was obtained. The overall result of the heuristic for a specific problem instance is the value of the best solution found for any value of $l$. Moreover, as computation time we provide the sum of the computation times spend for all applications for different value of $l$. \\

\begin{table}[!t]
\caption{Results of the heuristic. Each of the four subtables deals with one of the four problem instance sets.}
\label{tab:results:heuristic}
\centering
\subtable[Instances of \textsc{Group1}.]{
\label{tab:results:heuristic:group1}
\scalebox{0.75}{
\begin{tabular}{rrrrrr} \hline\hline
{\bf id} & $\;\;\;$ & \textsc{Best} & $\;\;\;$ & \multicolumn{2}{c}{Heuristic} \\ \cline{5-6}
         &          & \textsc{Known}         &          &  {\bf value}  &  {\bf time (s)} \\ \hline
1          && \ccg42   && \ccg42   &     7   \\
2          && \ccg48   && \ccg48   &     30  \\
3          && 55   && \ccg53   &     60  \\
4          && \ccg43   && \ccg43   &     1   \\
5          && 41   && \ccg$^+$40   &     15  \\
6          && 41   && \ccg$^+$40   &     6   \\
7          && 60   && \ccg57   &     22  \\
8          && 45   && \ccg44   &     1   \\
9          && \ccg43   && 46   &     6   \\
10         && \ccg58   && \ccg58   &     92  \\ \hline
{\bf avg.} && 47.7 && \ccg47.1 &	 24.0       \\
\hline\hline
\end{tabular}}}
\subtable[Instances of \textsc{Group2}.]{
\label{tab:results:heuristic:group2}
\scalebox{0.75}{
\begin{tabular}{rrrrrr} \hline\hline
{\bf id} & $\;\;\;$ & \textsc{Best} & $\;\;\;$ & \multicolumn{2}{c}{Heuristic} \\ \cline{5-6}
         &          & \textsc{Known}         &          &  {\bf value}  &  {\bf time (s)} \\ \hline
1          && 111   &&  \ccg103   &	 207    \\
2          && 114   &&  \ccg110   &	 214    \\
3          && 107   &&  \ccg99    &	 299    \\
4          && 111   &&  \ccg105   &	 261    \\
5          && 127   &&  \ccg120   &	 270    \\
6          && 102   &&  \ccg97    &	 252    \\
7          && 96    &&  \ccg91    &	 166    \\
8          && 114   &&  \ccg108   &	 223    \\
9          && 113   &&  \ccg109   &	 160    \\
10         && 98    &&  \ccg94    &	 158    \\ \hline
{\bf avg.} && 109.3 &&  \ccg103.6 &  221.0  \\
\hline\hline
\end{tabular}}}
\subtable[Instances of \textsc{Group3}.]{
\label{tab:results:heuristic:group3}
\scalebox{0.75}{
\begin{tabular}{rrrrrr} \hline\hline
{\bf id} & $\;\;\;$ & \textsc{Best} & $\;\;\;$ & \multicolumn{2}{c}{Heuristic} \\ \cline{5-6}
         &          & \textsc{Known}         &          &  {\bf value}  &  {\bf time (s)} \\ \hline
1          && \ccg171   &&   172   &    1030   \\
2          && 168   &&   \ccg165   &    671    \\
3          && 185   &&   \ccg180   &    1186   \\
4          && 179   &&   \ccg171   &    970    \\
5          && \ccg163   &&   \ccg163   &    610    \\
6          && 160   &&   \ccg155   &    660    \\
7          && 161   &&   \ccg160   &    242    \\
8          && 169   &&   \ccg166   &    276    \\
9          && \ccg169   &&   \ccg169   &    969    \\
10         && 161   &&   \ccg160   &    538    \\ \hline
{\bf avg.} && 168.6 &&   \ccg166.1 &    715.2  \\
\hline\hline
\end{tabular}}}
\subtable[Instances of \textsc{Real}.]{
\label{tab:results:heuristic:real}
\scalebox{0.75}{
\begin{tabular}{rrrrrr} \hline\hline
{\bf id} & $\;\;\;$ & \textsc{Best} & $\;\;\;$ & \multicolumn{2}{c}{Heuristic} \\ \cline{5-6}
         &          & \textsc{Known}         &          &  {\bf value}  &  {\bf time (s)} \\ \hline
1          && 86    &&  \ccg80    &	 142  \\
2          && 154   &&  \ccg144   &	 680  \\
3          && 113   &&  \ccg112   &	 232  \\
4          && 158   &&  \ccg157   &	 667  \\
5          && 165   &&  \ccg161   &	 499  \\
6          && 143   &&  \ccg139   &	 453  \\
7          && 131   &&  \ccg126   &	 482  \\
8          && 128   &&  \ccg120   &	 500  \\
9          && 142   &&  \ccg$^+$131   &	 437  \\
10         && 144   &&  \ccg136   &	 542  \\
11         && 121   &&  \ccg117   &	 412  \\
12         && 137   &&  \ccg130   &	 355  \\
13         && 171   &&  \ccg163   &	 1165 \\
14         && 146   &&  \ccg142   &	 530  \\
15         && 152   &&  \ccg145   &	 456  \\ \hline
{\bf avg.} && 139.4 &&  \ccg133.5 &  503.0  \\
\hline\hline
\end{tabular}}}
\end{table}

The results are shown in Table~\ref{tab:results:heuristic}, which contains one subtable for each of the four instance sets. Each subtable has the following format. The first column provides the instance identifier. The second column contains the value of the best solution found in the literature. Finally, the last two table columns present the results of our heuristic. The first one of these columns contains the value of the best solution generated by the heuristic, while the second column provides the total computation time (in seconds). The last row of each subtable presents averages over the whole subtable. Moreover, the best result for each instance is marked by a grey background, and those cases in which the result of applying CPLEX to $\MIPorig$ could be matched are marked by a "+" symbol.

The results allow to make the following observations. First, our heuristic is able to improve the best-known result from the literature in 37 out of 46 cases. In further six cases the best-known results from the literature are matched. Finally, in two remaining cases the heuristic is not able to produce a solution that is at least as good as the best-known solution known from the literature. Overall, the heuristic improves by $3.4\%$ (on average) over the best known results from the literature. On the downside, the heuristic is only able to match the results of applying CPLEX to model $\MIPorig$ in three out of 45 cases. However, this changes with growing instance size, as we will show later in Section~\ref{sec:large}.

\subsection{Gaining Insight into the Behavior of the Heuristic}

\begin{figure}[p]
\centering
\subfigure[Instance 10 of \textsc{Group1}, results (left) and computation time (right)]{
\label{fig:results:heuristic:a}
\includegraphics[width=0.42\textwidth]{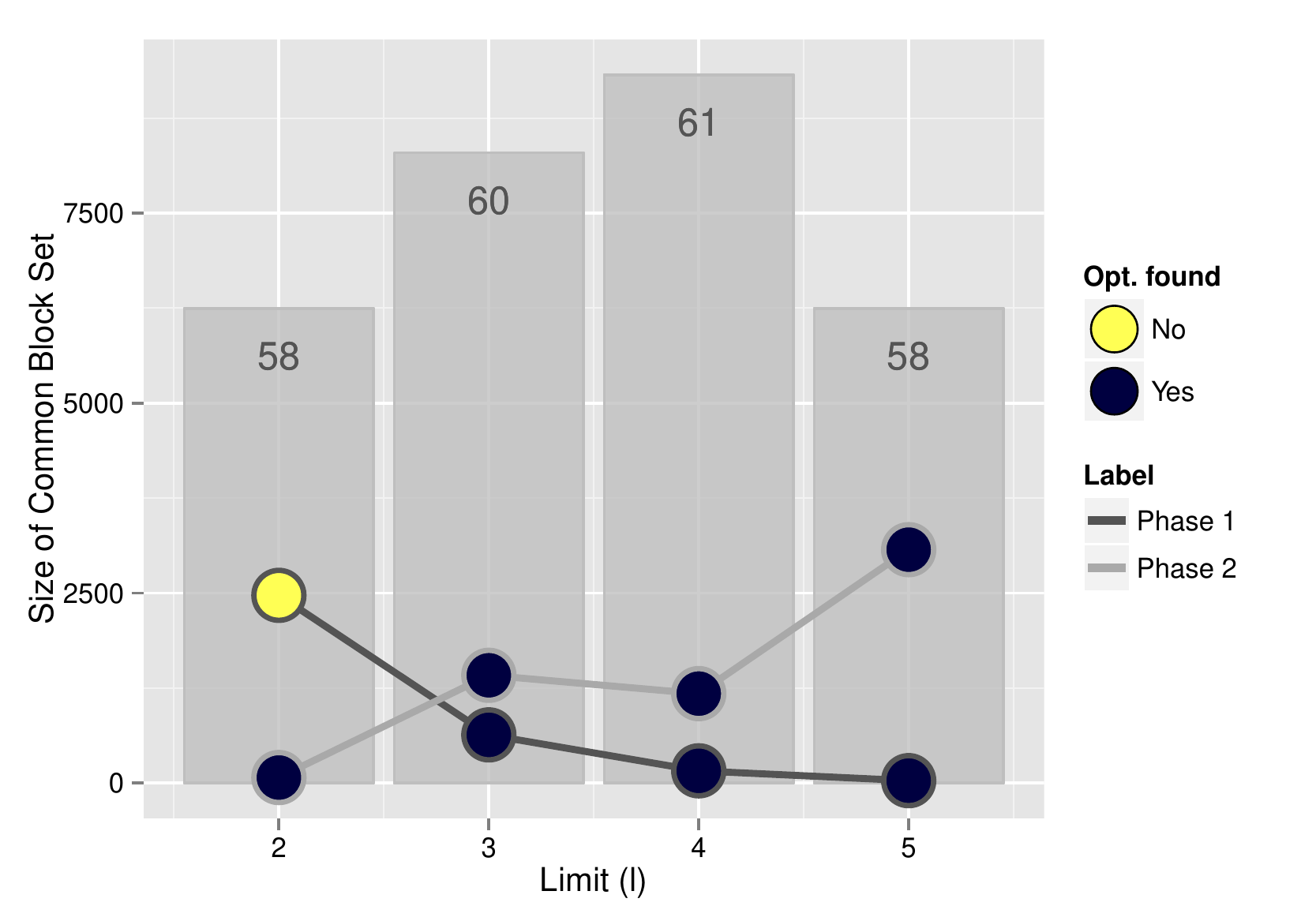}
\includegraphics[width=0.42\textwidth]{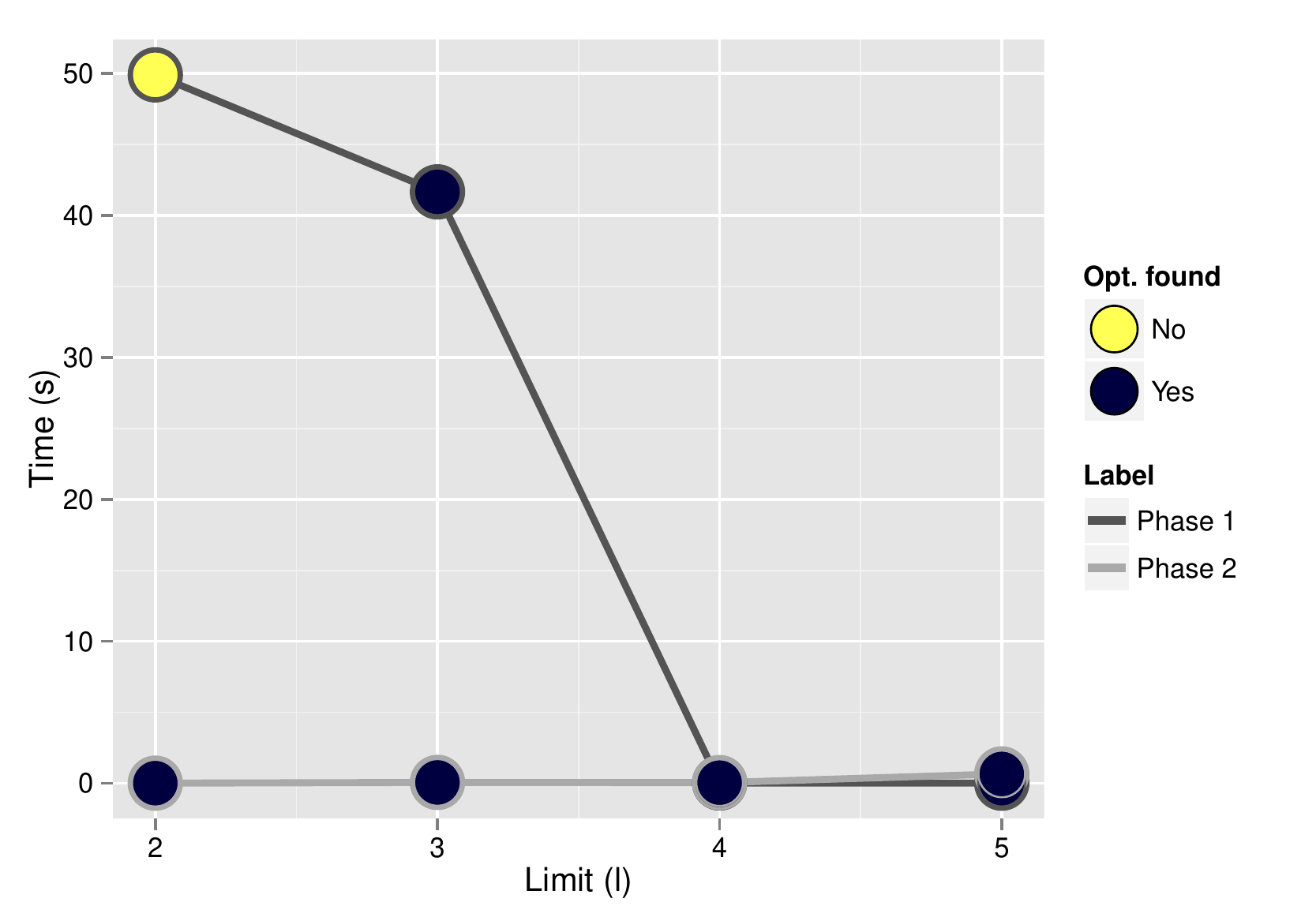}
}
\subfigure[Instance 5 of \textsc{Group2}, results (left) and computation time (right)]{
\label{fig:results:heuristic:b}
\includegraphics[width=0.42\textwidth]{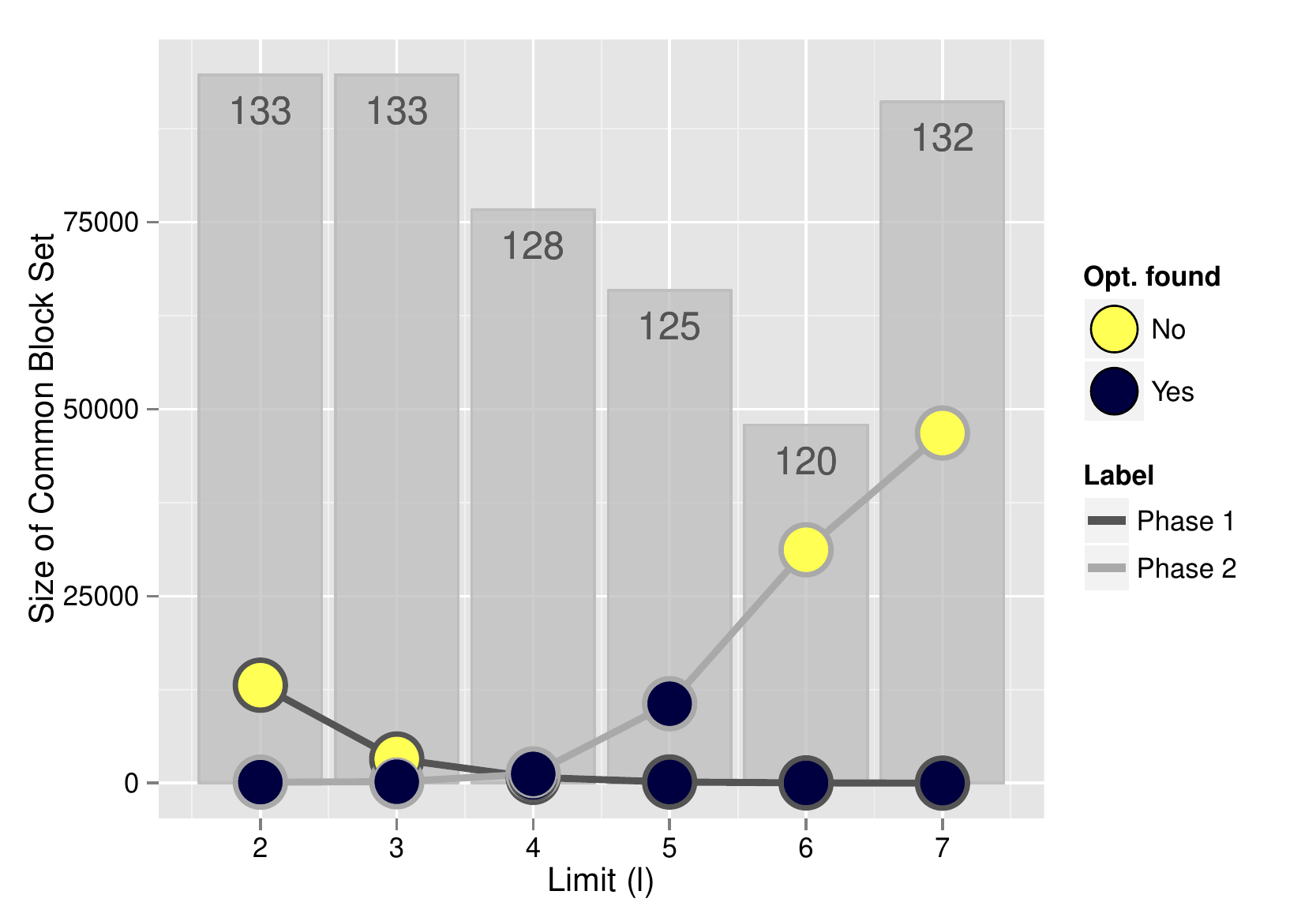}
\includegraphics[width=0.42\textwidth]{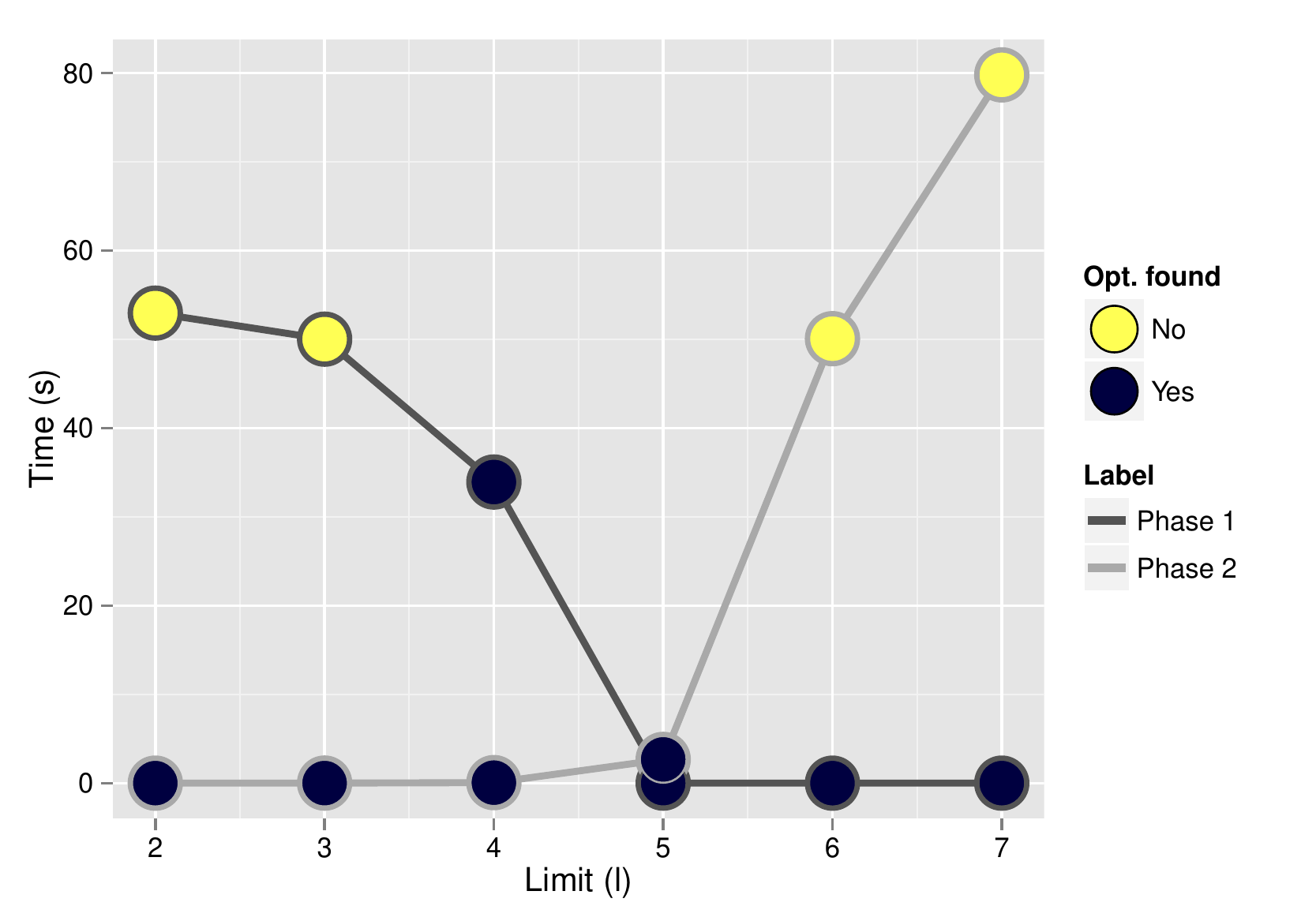}
}
\subfigure[Instance 3 of \textsc{Group3}, results (left) and computation time (right)]{
\label{fig:results:heuristic:c}
\includegraphics[width=0.42\textwidth]{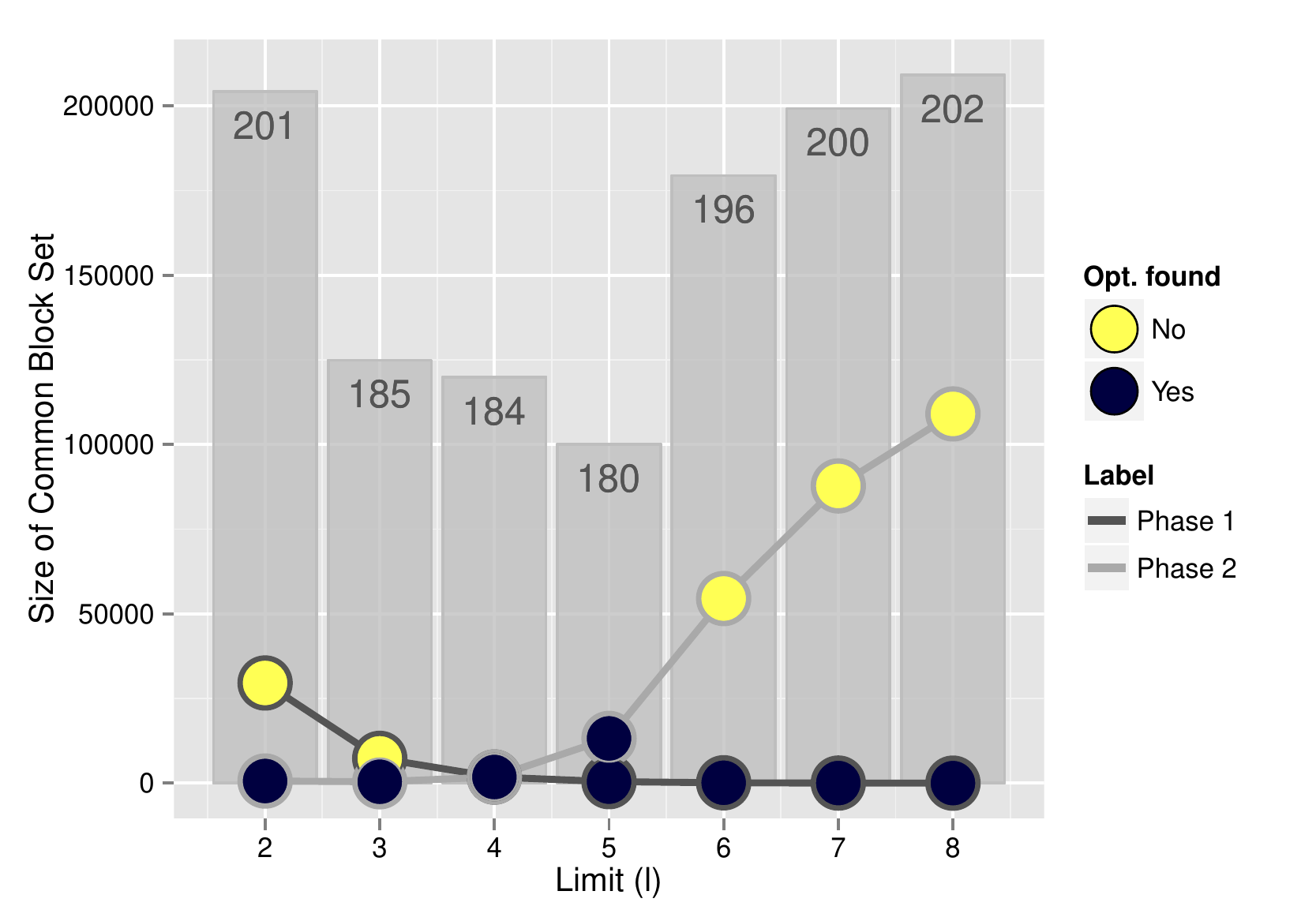}
\includegraphics[width=0.42\textwidth]{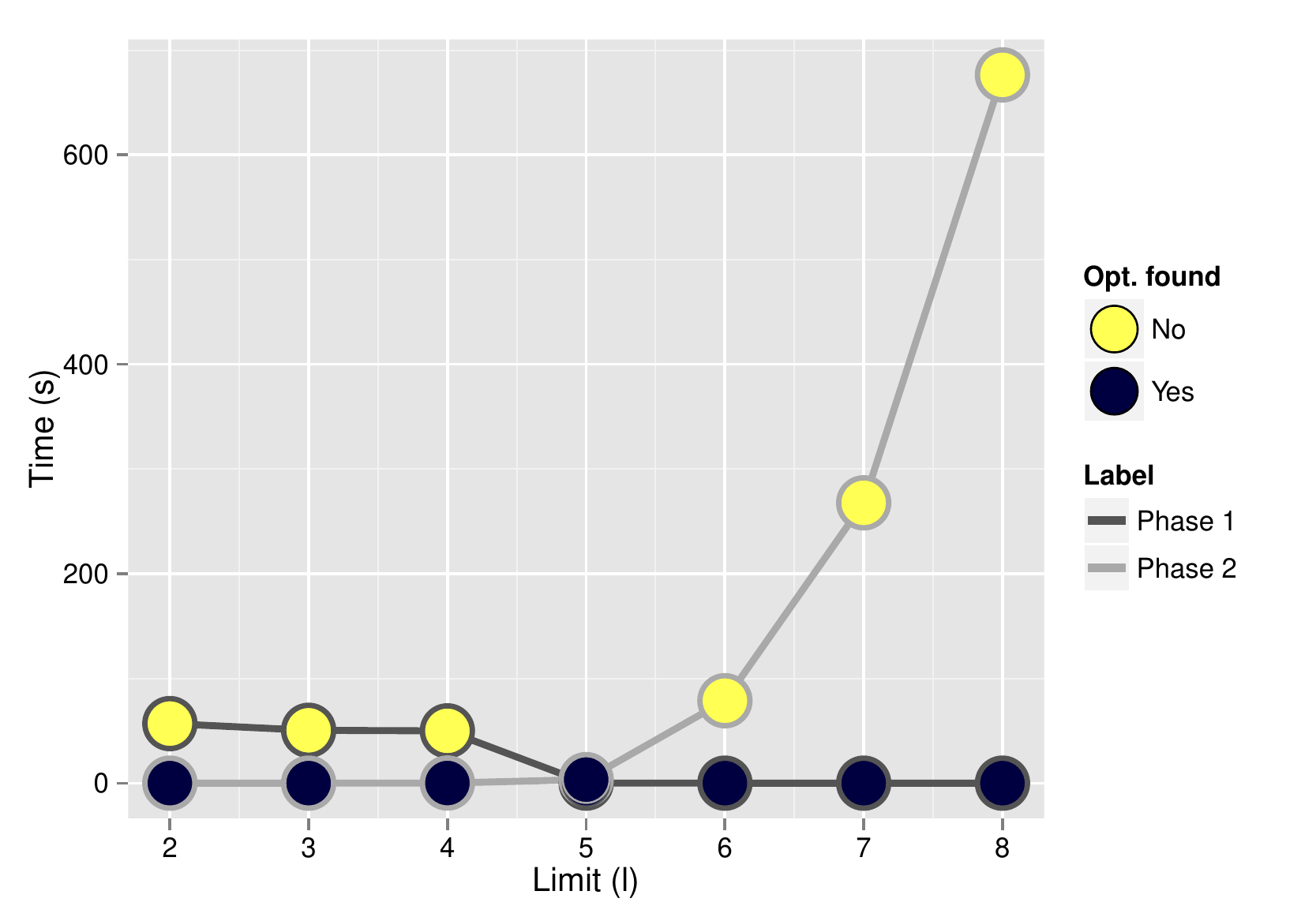}
}
\subfigure[Instance 13 of \textsc{Real}, results (left) and computation time (right)]{
\label{fig:results:heuristic:d}
\includegraphics[width=0.42\textwidth]{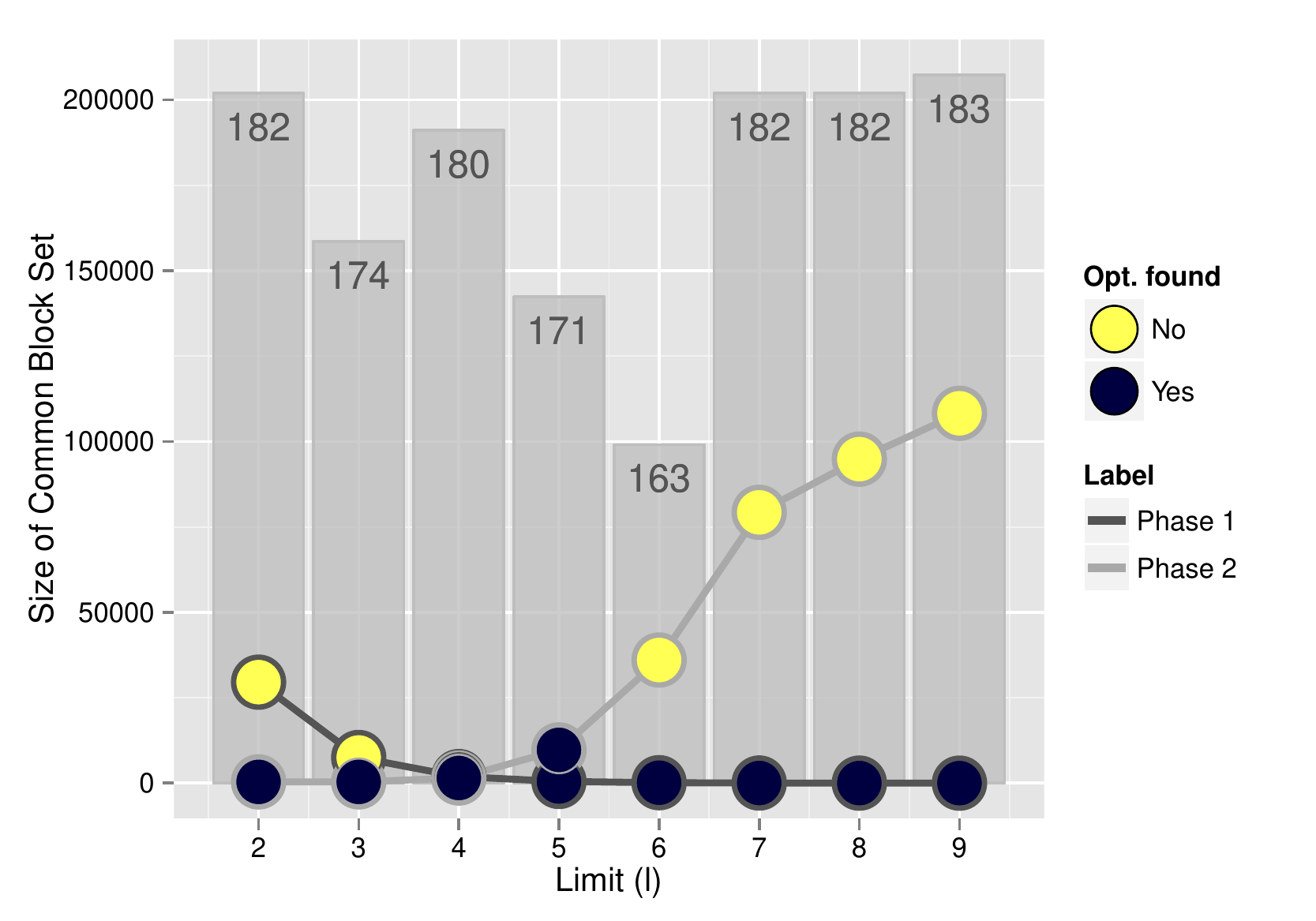}
\includegraphics[width=0.42\textwidth]{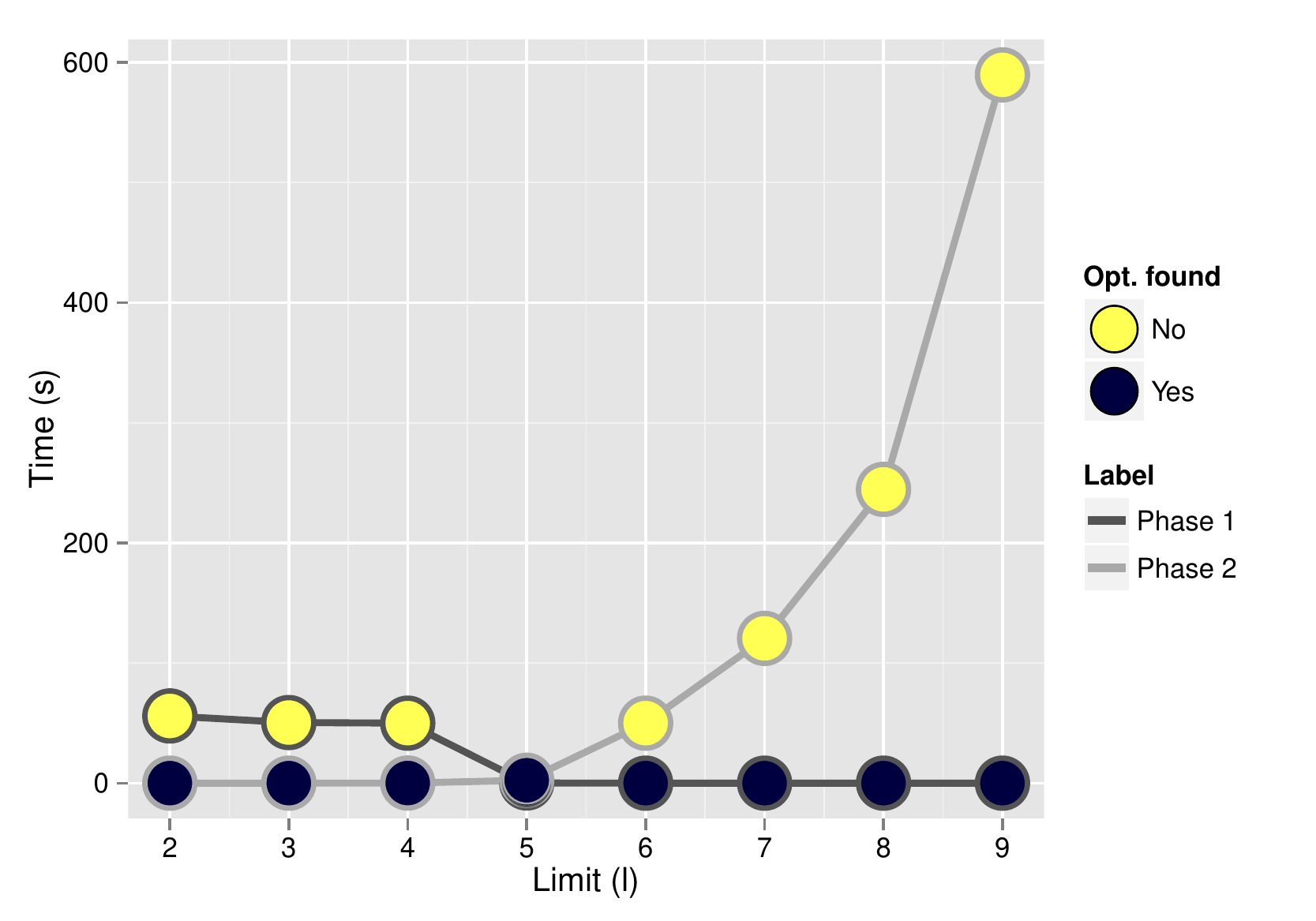}
}
\caption{Detailed information on the results of the proposed heuristic for four chosen problem instances. The description of the information content of the graphics is provided in the text.}
\label{fig:results:heuristic}
\end{figure}

With the aim of gaining more insight into the behavior of the heuristic with respect to the choice of a value for parameter $l$, the following information is presented in graphical form in Figure~\ref{fig:results:heuristic}. Two graphics are shown for each of the four chosen problem instances. More precisely, we chose to present information for the largest problem instances from each of the four instance sets (see subfigures (a) to (d) of Figure~\ref{fig:results:heuristic}). The left graphic of each subfigure has to be read as follows. The $x$-axis ranges over the possible values for $l$, while the $y$-axis indicates the size of the set of common blocks that is used for solving models $\MIPphone$ and $\MIPphtwo$. The graphic shows two curves. The one with a black line concerns solving model $\MIPphone$ in phase 1 of the heuristic, while the other one (shown by means of a grey line) concerns solving model $\MIPphtwo$ in phase two of the heuristic. The dots indicate for each value of $l$ the size of the set of common blocks used by the corresponding models. Moreover, in case the interior of a dot is light-grey (yellow in the online version) this means that the corresponding model could not be solved to optimality within 50 CPU seconds, while a black interior of a dot indicates that the corresponding model was solved to optimality. Finally, the bars in the background of the graphic present the values of the solutions that were generated with different values of $l$. The graphics on the right hand side present the corresponding computation times required by solving the different models. \\

The following observations can be made. When the value of $l$ is close to the lower or the upper bound---that is, either close to 2 or close to $l_{\mbox{\tiny max}}$---one of the two involved sets of common blocks is quite large, and, therefore, the computation time needed for solving the corresponding ILP may be large, in particular when the input instance is rather large. On the contrary, for intermediate values of $l$, the size of both involved sets of common blocks is moderate, and, therefore, CPLEX is rather fast in providing solutions, even if the optimal solution is not found (or can not be proven) within 50 CPU seconds. Moreover, the best results are usually obtained for intermediate values of $l$. This is with the exception of instance 10 of \textsc{Group1}, which might be an anomaly caused by the rather small size of the problem instance.

\subsection{Results of Heuristic for Larger Instances}
\label{sec:large}

Based on the findings of the previous subsection the heuristic was applied with an intermediate value of $l=5$ to all problem instances from the set of larger instances described at the end of Section~\ref{sec:mip:results}. The results are shown in Table~\ref{tab:results:heuristic:large}. The first table column provides the length of the input strings of the corresponding random instance. The second column indicates the result of applying CPLEX with a computation time limit of 3600 CPU seconds to $\MIPorig$.\footnote{Remember that the results of applying CPLEX to $\MIPorig$ were described in detail in Section~\ref{sec:mip:results}.} The remaining five columns contain the results of heuristic. The first one of these columns provides the value of the solution generated by the heuristic, while the second column shows the corresponding computation time. The next two columns provide the size of the sets of common blocks used in phase 1, respectively phase 2, of the heuristic. Finally, the last column gives information about the number of common blocks considered by the heuristic in comparison to the size of the complete set of common blocks (which can be found in Table~\ref{tab:results:mip:large}). In particular, summing the common block set sizes from phases 1 and 2  of the heuristic and comparing this number with the size of the complete set of common blocks, the percentage of the common blocks considered by the heuristic can easily be calculated. This percentage is given in the last table column. As in all tables of this paper, the best result per table row is marked by a grey background.

The following observations can be made. First, apart from the smallest problem instance, the heuristic outperforms the application of CPLEX to model $\MIPorig$. Moreover, this is achieved in a fraction of the time needed by CPLEX. Finally, it is reasonable to assume that the success of the heuristic is due to an important reduction of the common blocks that are considered (see last table column). In general, the heuristic only considers between $3.3\%$ and $6.5\%$ of all common blocks. This is why the computation times are rather low in comparison to CPLEX.

\begin{table}[!t]
\caption{Results of applying the heuristic in the context of larger instances.}
\label{tab:results:heuristic:large}
\centering
\scalebox{0.8}{
\begin{tabular}{rrrrrrrrr} \hline\hline
{\bf length} & $\;\;\;$ & CPLEX & $\;\;\;$ & \multicolumn{5}{c}{Heuristic} \\ \cline{3-3} \cline{5-9}
             &          & {\bf value}    &          & {\bf value} & {\bf time (s)} & {\bf $|B_{\geq 5}|$} & {\bf $|B_{ph2}|$}  & {\bf $\%$ of $B$} \\ \hline
800  && \ccg210  &&    225      & 16  &  801   & 13138  & $6.5\%$ \\
1000 && 304      &&    \ccg279  & 54  &  1385  & 12714  & $4.2\%$ \\
1200 && 342      &&    \ccg326  & 70  &  1785  & 27852  & $6.2\%$ \\
1400 && 401      &&    \ccg378  & 61  &  2535  & 20628  & $3.5\%$ \\
1600 && 442      &&    \ccg413  & 65  &  3244  & 24708  & $3.3\%$ \\
1800 && 486      &&    \ccg473  & 99  &  4416  & 44139  & $4.5\%$ \\
2000 && n.a.     &&    \ccg518  & 126 &  5132  & 61731  & $5.0\%$ \\
\hline\hline
\end{tabular}}
\end{table}

\section{Conclusions and Future Work}
\label{sec:conclusions}

In this paper we considered a problem with applications in bioinformatics known as the minimum common string partition problem. First, we introduced an integer linear programming model for this problem. By applying the IBM ILOG CPLEX solver to this model we were able to improve all best-known solutions from the literature for a problem instance set consisting of 45 instances of different sizes. The smallest ones of these problem instances could even be solved to optimality in very short computation time. The second contribution of the paper concerned a 2-phase heuristic which is strongly based on the developed integer linear programming model. The results have shown that, first, the heuristic outperforms competitor algorithms from the literature, and second, that it is applicable to larger problem instances.

Concerning future work, we aim at studying the incorporation of mathematical programming strategies based on the introduced integer linear programming model into metaheuristic techniques such as GRASP and iterated greedy algorithms. Moreover, we aim at identifying other string-based optimization problems for which a 2-phase strategy such as the one introduced in this paper might work well.

\paragraph*{\bf Acknowledgements} C.~Blum was supported by project TIN2012-37930 of the Spanish Government. In addition, support is acknowledged from IKERBASQUE (Basque Foundation for Science). J.~A.~Lozano was partially supported by the Saiotek and IT609-13 programs (Basque Government), TIN2010-14931 (Spanish Ministry of Science and Innovation), COMBIOMED network in computational bio-medicine (Carlos III Health Institute)

%
%

\end{document}